%% file: main.tex
\documentclass[hidelinks, lettersize,journal]{IEEEtran} 

\IEEEoverridecommandlockouts

\usepackage[dvipsnames]{xcolor}
\usepackage[ruled,vlined]{algorithm2e}
\usepackage{algpseudocode}
\usepackage{amsmath}

\usepackage{enumitem}
\usepackage{amssymb}
\usepackage{epsfig}
\usepackage{booktabs}
\usepackage{float}
\usepackage{url}
\usepackage{gensymb}
\usepackage{multirow}
\usepackage{tabularx}
\usepackage{pifont}
\usepackage{amsmath}
\usepackage{siunitx}
\usepackage{hyperref}
\hypersetup{
    colorlinks=false,
    urlcolor=red,
    }

\usepackage[backend=biber,
            hyperref=true,
            url=false,
            isbn=false,
            doi=false,
            backref=false,
            style=ieee,
            citestyle=numeric-comp,
            sorting=none,
            block=none]{biblatex}

\DeclareSIUnit{\inchHtwoO}{\text{inH\textsubscript{2}O}}

\usepackage{array,multirow,graphicx}
\usepackage{color}
\usepackage{wrapfig}
\newcommand{\ourmethod}{FORTE}

\newcommand{\webpage}{{\tt\small\textbf{\textcolor{orange}{\url{https://merge-lab.github.io/FORTE/}}}}}

\title{\LARGE \bf \ourmethod{}: Tactile Force and Slip Sensing on Compliant Fingers for Delicate Manipulation}

\author{
Siqi Shang, Mingyo Seo, Yuke Zhu, and Lillian Chin
\thanks{This work was partially supported by the Texas Robotics Industrial Affiliate Program, the National Science Foundation (FRR-2145283, EFRI-2318065), the Office of Naval Research (N00014-24-1-2550), the DARPA TIAMAT program (HR0011-24-9-0428), and the Institute of Information \& Communications Technology Planning \& Evaluation (IITP) grant funded by the Korean Government (MSIT) (No. RS-2024-00457882, National AI Research Lab Project). All authors are with The University of Texas at Austin, TX, USA. Correspondence: {\tt\footnotesize siqi.shang@utexas.edu}}
}

\begin{document}

\maketitle
\input{sections/abstract}
\input{sections/introduction}
\input{sections/related_work}
\input{sections/method}
\input{sections/characterization}
\input{sections/results}
\input{sections/conclusion}

\renewcommand*{\bibfont}{\scriptsize}
\begingroup
\printbibliography
\endgroup

\end{document}

%% file: sections/abstract.tex
\begin{abstract}
Handling fragile objects remains a major challenge for robotic manipulation. Tactile sensing and soft robotics can improve delicate object handling, but typically involve high integration complexity or slow response times.
We address these issues through \ourmethod{}, an easy-to-fabricate tactile sensing system. \ourmethod{} uses 3D-printed fin-ray grippers with internal air channels to provide low-latency force and slip feedback. This feedback allows us to apply just enough force to grasp objects without damaging them. We accurately estimate grasping forces from 0--8~\si{\newton} with an average error of 0.2 \si{\newton}, and detect slip events within 100 \si{\milli\second} of occurring. \ourmethod{} can grasp a wide range of slippery, fragile, and deformable objects, including raspberries and potato chips with 92\% success and achieves 93\% accuracy in detecting slip events. These results highlight \ourmethod{}’s potential as a robust solution for delicate robotic manipulation. Project page: \webpage{}
\end{abstract}

\vspace{-10pt}

%% file: sections/introduction.tex
\section{Introduction} 

\IEEEPARstart{H}{umans} delicately grasp objects by applying \textit{just enough} force. Apply too little force and the object will slip, but apply too much force and the object will be damaged~\cite{wheat2004humanfragileforce}. Humans navigate this tradeoff through the real-time force and slip feedback provided by mechanoreceptors embedded within their skin \cite{westling1987responsesmechanoreceptor,  johansson2009codinghumanskinforobjectmanipulation}. However, current robots lack comparable human-like sensing capabilities in their fingers.  
% Current robots do not have comparable force and slip sensors in their fingers.
% Current robotic solutions to achieving human-like sensing capabilities suffer from several limitations.
Most parallel grippers struggle with fragile objects due to their sole reliance on visual feedback and simplified open-close gripper actions \cite{seo2024legato, zhao2023learning, chi2024universal}. There have been two main approaches to improve these grippers: (1) adding compliance, and (2) incorporating tactile sensing. In the former approach, researchers adopt designs like the fin-ray gripper \cite{crooks2016fin} so binary open-close actions deform the gripper rather than the object. However, this passive deformation makes precise force control difficult, making it challenging to handle objects with varying fragility \cite{crooks2016fin, della2020dexteritysoftfinger}. In the latter approach, researchers add tactile sensors to existing grippers \cite{romeo2020methodssurvey, chen2018tactilesurvey}. While effective, these sensors tend to either be expensive off-the-shelf sensors with durability issues \cite{liu2023gelsight, wang2024gelsightflexiraybreakingplanar, zhang2022vision} or custom sensor technologies with complex fabrication or signal processing needs \cite{shih2017customresistiveskin, hashizume2019capacitivegecko, qi2023pvdf, diao2025triboCus}. To achieve delicate robotic manipulation, we need simple compliant end-effectors that incorporate force and slip sensing.

\begin{figure}[t]
	\centering
	\includegraphics[width=0.98\linewidth]{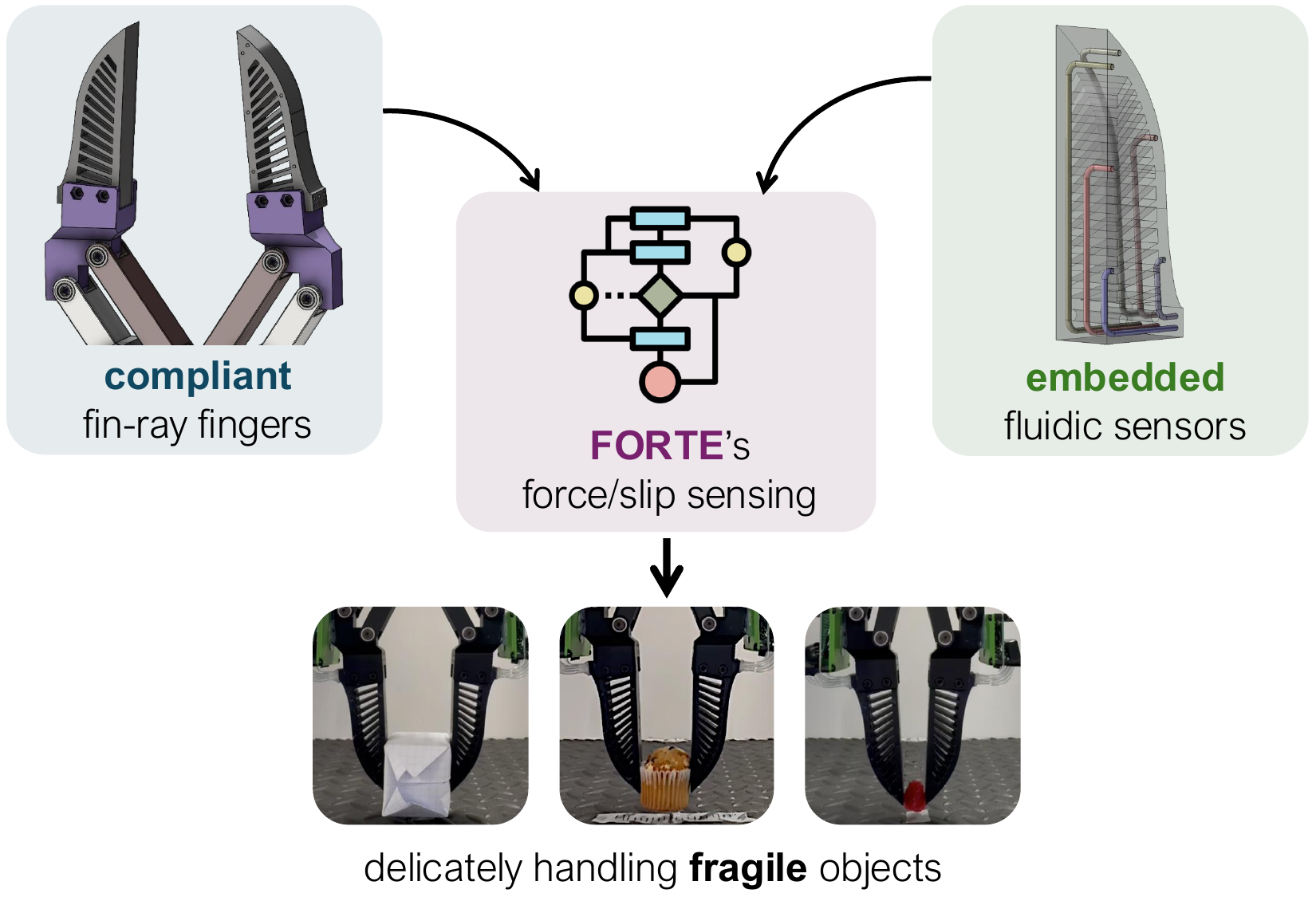}
        \vspace{-7pt}
        \caption{\textbf{Overview.} \ourmethod{} enables the delicate manipulation of fragile objects by leveraging passive compliance and tactile feedback. This is achieved by integrating compliant fin-ray fingers and embedded fluidic sensors with algorithms for force estimation and slip detection.}
        \label{fig:header}
     \vspace{-12pt}
\end{figure}

In this paper, we introduce \ourmethod{} (\underline{F}ragile \underline{O}bject G\underline{r}asping with \underline{T}actile S\underline{e}nsing), a tactile force and slip sensing system embedded in compliant fingers for delicate manipulation. Akin to human grasping, \ourmethod{} applies just the right amount of force to grasp delicate objects and dynamically adds more force if the object slips. We do so by combining the core technologies of fin-ray-style fingers and fluidic innervation (Fig.~\ref{fig:header}). Both technologies are easily manufacturable through 3D-printing, making \ourmethod{} easy to build. Instead of mounting external tactile sensors, \ourmethod{} leverages internal air channels embedded within the compliant finger structure for sensing. The channels deform as the finger deforms, providing a pressure signal that can be read by off-the-shelf transducers. Since the signals directly follow from the gripper's internal deformation, \ourmethod{} provides precise force estimation (0-8~\si{\newton} $\pm$ 0.19~\si{\newton}). The sensors' fast response and high temporal resolution also enable a frequency-based slip detection between the gripper and a grasped object. \ourmethod{} enables the robot to grasp a wide range of fragile, slippery, and deformable objects with a 91.9\% success rate across 310 trials. We make the following contributions in this work:
\begin{itemize}
    \item We design and manufacture compliant fingers with internal air channels for sensing. The fingers sense contact events with a high sampling rate of 2 \si{\kilo\hertz}.
    \item We achieve accurate force estimation with validation RMSE below 0.2 \si{\newton}.
    \item We develop a novel analytical slip detection algorithm that achieves 0.91 F1 score on 31 items with 29 unseen.
    \item We demonstrate \ourmethod{}’s capability for precise force estimation and slip detection through single-trial grasps of 31 different items, achieving a 91.9\% success rate.
    \item We publicly release the hardware designs and algorithmic implementations on the project webpage.
\end{itemize}

%% file: sections/related_work.tex
\section{Background}
\label{sec:relatedWork}

\subsection{Compliant Grippers with Tactile Feedback}
\label{sec:related_grippers}
To incorporate tactile sensing in a compliant gripper, the sensor must \textit{not} introduce unwanted stiffness. For example, a sensor with a rigid backing creates a stiff patch, compromising the finger’s compliance. Vision-based tactile sensing offers one non-intrusive way to preserve structural compliance, by pairing a soft elastomeric surface with an internal camera. The camera tracks the surface's deformation, enabling high-resolution measurement of contact \cite{liu2022gelsight, wang2024gelsightflexiraybreakingplanar, zhang2022vision}. This large spatial resolution comes at the cost of added structural complexity from the camera and durability issues from the elastomer. The elastomeric surface is often \textit{too} compliant and tears under repeated or high-force interactions \cite{zhao2025polytouch}.

Acoustic tactile sensors also preserve structural compliance, as microphones can be embedded underneath the contact surface of the finger. The high-frequency acoustic signals captured by a microphone enable contact force and location sensing \cite{zoller2020active}, material identification \cite{ wall2023pasacacousticsoft}, and contact-rich policy learning \cite{liu2024maniwav}. However, their performance is limited by environmental noise and the complexity of interpreting signals under compliant contact.

Finally, barometric solutions work by adding internal air chambers to the soft fingers. Since air is a compressible fluid, these empty air chambers will match the fingers' softness and report a measurable pressure change as the fingers deform. Barometric tactile sensors can detect dynamic contact events such as object sliding, force sensing and simultaneous contact localization \cite{gruebele2021stretchablebaroskin, he2020softfingertip, choi2023integratedbarosg}. Barometric sensing avoids the durability challenges of vision-based sensors and eschews the complex signal processing requirements of acoustic-based sensors, but have also had their own complex fabrication challenges.

In this work, we apply fluidic innervation, a barometric sensing technique, to the fin-ray gripper. Fluidic innervation alleviates the fabrictation difficulties of traditional barometric sensing by 3D printing the finger structure with embedded air channels, resulting in easy fabrication and precise force estimation and slip detection capabilities~\cite{truby2022fluidic, zhang2024embedded}. 
% Building on the sensing principles in prior studies, this work adapts fluidic innervation to a fin-ray gripper geometry, enabling slip detection and demonstrating viability across a wide range of objects.

\begin{figure}[]
\centering
    \includegraphics[width=\linewidth]{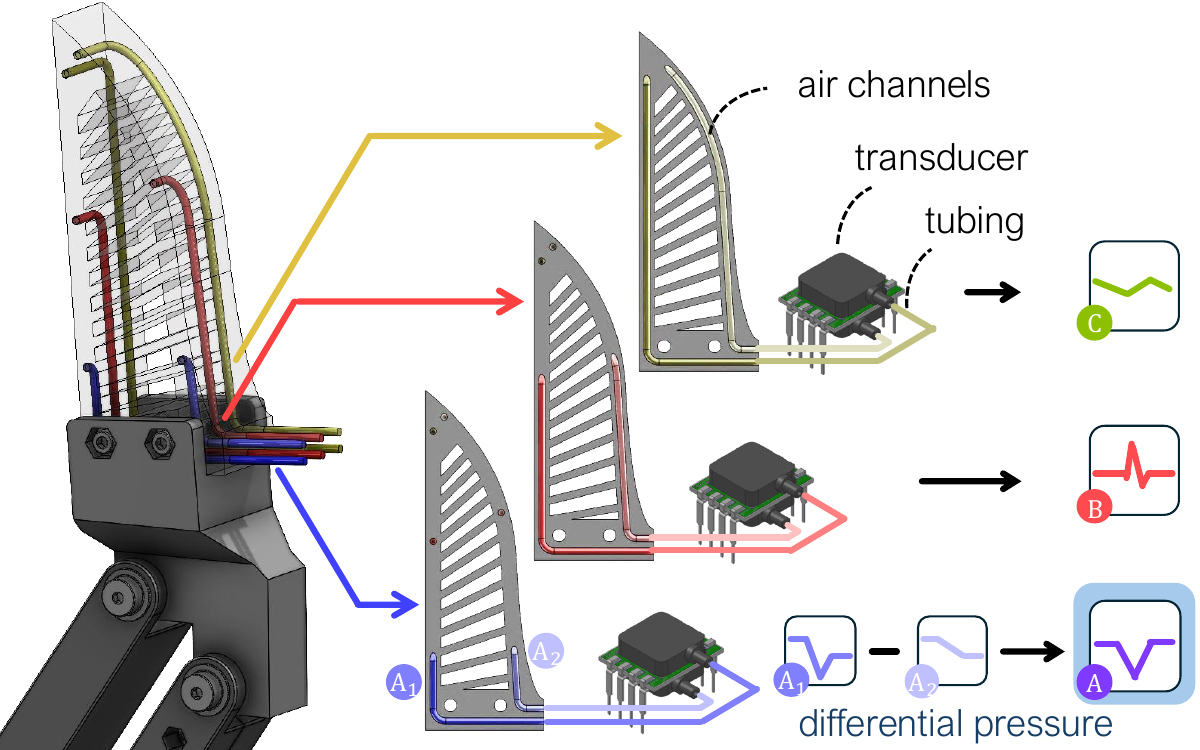}
    \vspace{-15pt}
	\caption{\textbf{Hardware Design of \ourmethod{}.} Fin-ray fingers with internal empty air channels are 3D printed as a single structure. The air channels act as a tactile sensor by measuring their pressure with an off-the-shelf pressure transducer. Each pair of channels (marked as the same color) under the inner and outer surfaces of the finger is sealed on the side surface and connected to one differential air pressure transducer.}
    \label{fig:design}
    \vspace{-12pt}
\end{figure}

\subsection{Slip Definition and Detection}
\label{sec:slipDefinition}
Slip events are broadly classified into \textbf{gross slip} and \textbf{incipient slip} \cite{howe1989slipdef, chen2018tactilesurvey}. Gross slip refers to relative motion occurring across the entire contact interface. Incipient slip describes the partial slip that initiates within the contact area while the rest remains stationary. A slip event typically starts with incipient slip, when external forces overcome local static friction to cause motion. In a parallel gripper, if a slip appears only at one finger, this is also considered incipient. In addition to these general categories, slip events can also be classified by the type of relative motion between the contacting surfaces:

\begin{itemize}
    \item \textbf{Translational:}  relative linear motion between contacting surfaces along the tangential direction.
    \item \textbf{Rotational:}  relative angular motion between contacting surfaces around the normal axis.
    \item \textbf{Stick-Slip:} the alternating motion between static friction contact and sudden slipping \cite{byerlee1970mechanicsstickslip}.
\end{itemize}

To identify slip events during grasping, two major methods have emerged: (1) detecting relative motion at the contact surface and (2) sensing vibrations. Method (1) is commonly achieved by using vision-based tactile sensors to track the displacement of contact textures or embedded markers. These approaches can detect translational and rotational incipient slip with high spatial precision \cite{dong2019maintaining, li2018sliptabletop, funk2024evetac}, but sometimes require a large normal force to produce a well-defined contact profile (ex. 50  \si{\newton} for a 132 g crayon box) \cite{dong2019maintaining}. This level of force is infeasible for delicate manipulation.

Method (2) works by identifying frequency domain features caused by micro-movements at the contact interface, such as the power peak within frequency bands  \cite{kim2022barotac, li2022multifrequencystickslip, romano2011human, fernandez2014microvibriation}. These methods are difficult to generalize because of a low signal-to-noise ratio and the highly variable sensor response from different combinations of contact surface properties~\cite{chen2018tactilesurvey}. More specific features can be chosen, such as the magnitude of the tangential force derivative \cite{su2015biotacslowopen}, but require accurate and timely force sensing. Rather than select a specific feature, learning-based methods directly operate on raw sensor data \cite{schopfer2010using, meier2016tactileCNN, veiga2015stabilizing, judd2022slip, grover2022learning}, at the cost of collecting high-fidelity slip data at scale. 

We address these issues by selecting a second-order feature that is robust to noise and requires minimal normal force to achieve. Our method can effectively capture stick-slip events without additional algorithmic dependencies or special hardware. Unlike high-dimensional vision-based methods that often rely on data-driven models, our approach directly measures finger deformation with higher sensitivity, creating an analytically tractable slip detection algorithm.

%% file: sections/method.tex
\section{Methods}
\subsection{Hardware Design and Sensorization}
We design \ourmethod{}'s fingers based on the popular fin-ray fingers \cite{crooks2016fin}. When deformed by a grasped object, the fin-rays' geometric structure creates a tight conformation to the object, creating a secure grasp. However, this large deformation also makes it challenging to seamlessly incorporate tactile sensing. We sensorize fin-rays using fluidic innervation, a barometric tactile sensor created by 3D-printing structures with empty air channels inside \cite{truby2022fluidic, zhang2024embedded}. As the structure deforms, the air channels deform as well. This resulting change in air pressure can be measured with off-the-shelf pressure transducers. As fluidic innervation does not change a structure's overall geometry, it preserves the fin-ray's high deformation.

% We print the fin-rays out of EPU 40 on a Carbon M2 printer with air channels embedded under the inner and outer surface of the finger (Fig.~\ref{fig:design}), following the protocol in Truby et al.

The fin-rays are fabricated from EPU 40 on a Carbon M2 printer, following the material selection and fabrication procedure reported by Truby et al \cite{truby2022fluidic}. To summarize, we print a standard fin-ray finger, modified to have internal empty air channels under the inner and outer surface of the finger (Fig.~\ref{fig:design}). These air channels are cleaned out with solvent, sealed at one end and connected to a pressure transducer at the other. These sensors are \textit{passive} sensing elements; they only contain ambient air and are not pressurized or actuated. As the fin-ray structure deforms, the internal air pressure of the channels will change, capturing the deformation of the fin-ray under external contact. We design our channels in an L-shape to maximize vertical and horizontal information. Each channel has a diameter of 2 \si{\milli\meter}, with a 1 \si{\milli\meter}-thick wall between the channels and the adjacent surface. The distal ends of the channels are sealed on the finger's side surface with silicone epoxy (Sil-Poxy, Smooth-On). Each transducer (All Sensors ELVR-L01D, 1 \si{\inchHtwoO}) is connected to the corresponding inner/outer channels in the finger via silicone tubing. Since each pressure transducer measures the \textit{difference} in pressure, each transducer measures the \textit{net} pressure change between the two channels in the pair. 

Each finger is sensorized with 3 transducers, resulting in 6 sensor signals total across \ourmethod{}. The transducers are connected to the analog inputs of an ESP32-S3 and sampled at 2 \si{\kilo\hertz} at an 11-bit resolution to capture rapid and transient dynamics that occur over short time scales. The analog readings are normalized to the range of $-1$ to $1$ and transmitted at 2 \si{\kilo\hertz} to a host PC. The sensor signals are filtered with a median filter of size 11 to reduce impulsive noise while preserving transient features, inducing at most 2.5 \si{\milli\second} of latency.

\subsection{Force Estimation Algorithm}
\label{sec:f_e}
We estimate the gripping force directly from the filtered internal pressure readings. We employ Support Vector Regression (SVR) with a radial basis function (RBF) kernel. To account for sensor drift and stress relaxation, three additional features are introduced for each channel, representing the mean values of the most recent 2.5, 5, and 10 seconds of historical data.
Let \( S(t) \in \mathbb{R}^6 \) be the sensor signal frame at time $t$, the feature vector \(z(t) \) is defined as: 
\[ z(t) = \bar{S}_{2.5}(t) \oplus \bar{S}_{5}(t) \oplus \bar{S}_{10}(t) \] where \( \bar{S}_{\tau}(t) \in \mathbb{R}^6 \) denotes the mean of the past \( \tau \) seconds of sensor data.
The most recent sensor frame \( S(t)\) is concatenated with \( z(t)\) to form the input vector \( v = S(t) \oplus z(t) \in \mathbb{R}^{24} \). 

\begin{table}[t]
\centering
\caption{Slip Detection Parameters}
\vspace{-5pt}
{
\begin{tabular}{@{~}l@{~~}l@{~}}
\hline
\textbf{Parameter} & \textbf{Value} \\ \hline
Median filter window size & $M = 11$ \\
FFT window size & $N = 400$ \\
Overlap & $O = 0.99$ \\
Frequency band & $[f_{\text{min}}, f_{\text{max}}] = [10,50]$\,\si{\hertz} \\
PSD history length (moving window size) & $V = 15$ \\
Minimal increment for monotonicity & $\delta = 0.1$\,\si{\decibel} \\
Minimum group average variance & $\alpha = 0.6$\,\si{\decibel}$^2$ \\
Slip detection threshold & $T = 2$\,\si{\decibel}$^2$ \\
\hline
\end{tabular}
}
\label{tab:selected_parameters}
\vspace{-8pt}
\end{table}

\subsection{Slip Detection}
\label{sec:slipAlgo}

The high temporal resolution of the sensors means that we can detect object slip by tracking transient changes in the applied force over time and detect object slip. Our method tracks the increase in spectral power within a specific frequency band.

First, we apply a Discrete Fourier Transform to the sensor signal $S_i(t)$ for a given sensor index $i$. We then divide the frequency-domain signal into $K$ overlapping windows of length \(N\) with step \(N-O\cdot N\), where $O$ is the overlap factor. For each segment \(k\), the sensor \(i\)'s signal is first windowed and then transformed into the frequency domain to obtain 
\begin{equation}
\label{eq:psd}
X_{i,k}(f)=\sum_{n=0}^{N-1} x_i(n)w(n)e^{-j2\pi f n},
\end{equation}
where \( x_i(n) = S_i\left(t_k - \frac{N - 1 - n}{f_s} \right) \) with $n$ indexing the \( N \) samples of sengment $k$ ending at time \( t_k \), \( w(n) \) is the Hann window function, and \(j\) is $\sqrt{-1}$. Thus, the sensor signal of segment \(k\) for sensor \(i\) is implicitly present in \(X_{i,k}(f)\). 

The expression \(\left|X_{i,k}(f)\right|^2\) gives an estimate of the power at frequency \(f\) for that segment, and normalizing these values yields the power spectral density \(P_i(f)\) for sensor \(i\) in the unit of $(\si{\inchHtwoO})^2/\si{\hertz}$. For each sensor \(i\) and the most recent window \(x\), we compute the PSD \(P_i\) for normalized frequency $f_k = \frac{k}{N}, \ 0 \le k < N$ using Welch's method as:
\begin{equation}
    P_i(f_k) = \frac{1}{f_s\displaystyle\sum_{n=0}^{N-1}w^2(n)} \left|X_{i,k}(\frac{k}{N})\right|^2\,.
\end{equation}
For each sensor's computed PSD, we select the maximum value across its frequency bins to form a scalar feature. We define this PSD feature as:
\begin{equation}
P_i^{\max}=\max_{f\in [f_{\text{min}},f_{\text{max}}]}\Bigl\{10\log_{10}\bigl(P_i(f)+\epsilon\bigr)\Bigr\}\,.
\label{equation:PSD}
\end{equation}
We add a small constant \(\epsilon = 1 \times 10^{-12}\) to the power spectral density estimate before taking the logarithm to prevent the potential numerical instability of taking $\log 0$. The PSD feature represents the maximum value of power magnitude within the selected frequency range $[f_{\text{min}}, f_{\text{max}}]$.

\begin{figure*}[t]
    \centering
	\includegraphics[width=1.01\linewidth]{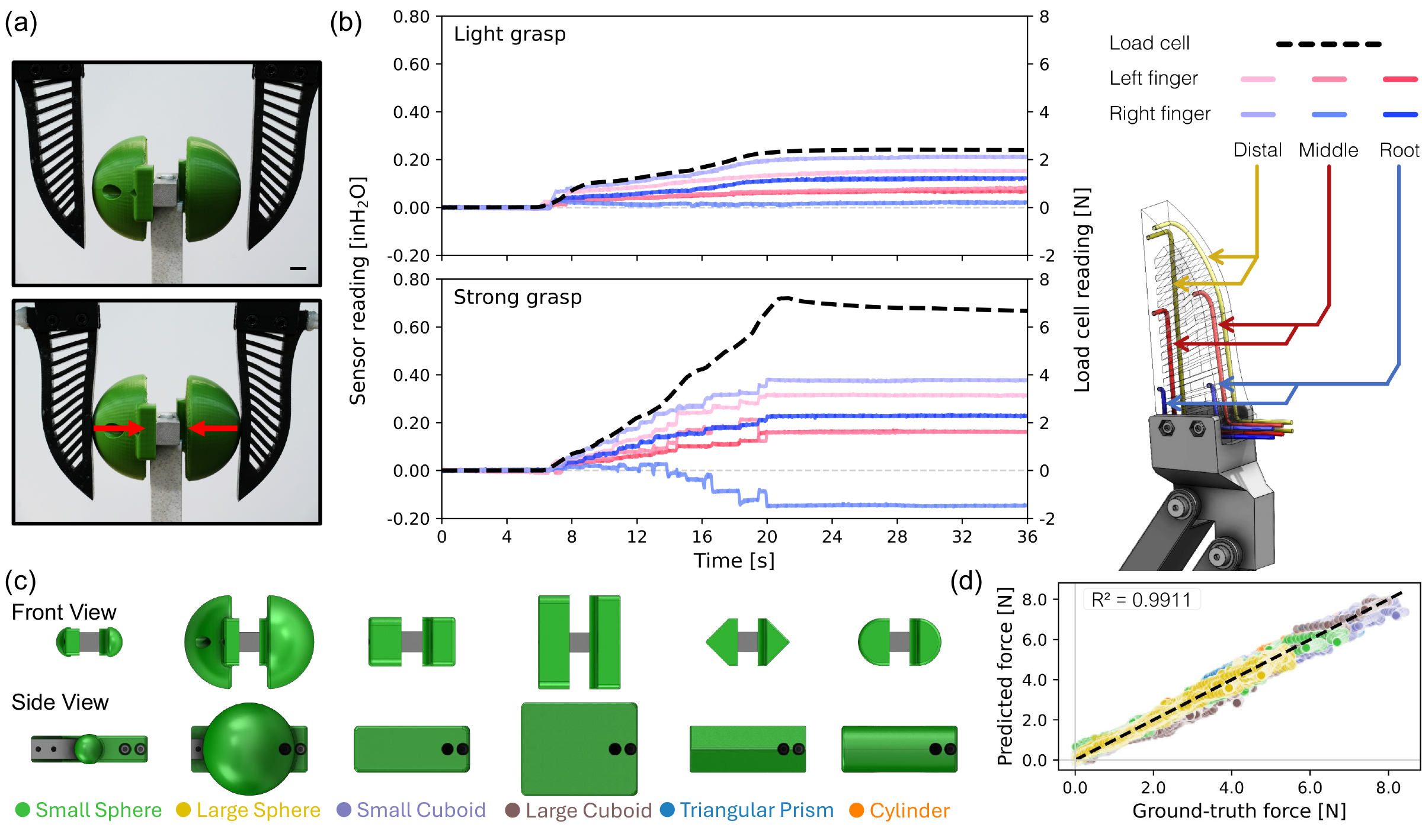}
        \vspace{-23pt}
        \caption{\textbf{Grip Force Estimation Characterization} (\textbf{a}) Snapshots of gripper in open and closed poses during data collection. Grip force of the fingers are measured with the horizontally placed loadcell testing rig. Scale bar represents 1~\si{\cm}. (\textbf{b}) Sensor and load cell readings over time for a light and strong grip. Red and blue lines show sensor readings from the left and right fingers, respectively, with each line corresponding to a channel pair at distal, middle, or root sensing location. Sensor outputs exhibit a consistent monotonic trend with increasing grip force, with the distal sensors showing the largest response. This trend confirms that the pressure signals scale with finger deformation and support accurate force estimation. (\textbf{c}) Renders of the custom load-cell testing rigs. (\textbf{d}) Predicted force versus ground-truth force for models trained with the feature vectors from Section~\ref{sec:f_e} as input. Colored dots represent data points collected using the corresponding testing rig, while the dashed black line indicates perfect prediction.}
    \label{fig:force_characterization}
    \vspace{-5pt}
\end{figure*}

To quantify temporal fluctuations in spectral power and search for our large-magnitude frequency feature, we compute the moving variance of the PSD magnitude as a second-order statistical feature. For each sensor $i$, we create a double-ended queue of \(P_i^{\max}\) as a history $H_i$, which has a fixed length $V$. We then compute the moving variance over the feature history:
\begin{equation}
   \sigma_i^2=
\begin{cases}
\operatorname{Var}(H_i), & \text{if } H_i(k+1)-H_i(k)>\delta,\ \forall\, k,\\[1ex]
0, & \text{otherwise\,.}
\end{cases}
\end{equation}

Each sensor $i$ belongs to either the right finger or the left finger. To detect slip at the finger level, we average the moving variance across the entire finger, which also helps reduce the impact of sensor noise. Let the finger group $G$ denote the set $\{R, L\}$ for the right and left fingers, respectively. The average moving variance is:
\begin{equation}
  \bar{\sigma}_g =
\begin{cases}
\displaystyle \frac{1}{|g|}\sum_{i\in g}\sigma_i^2, & 
\text{if }\bar{\sigma}_g \geq \alpha,\ \forall g \in G, \\[1ex]
0, & \text{otherwise\,.}
\end{cases}\,  
\label{eq:avgMV}
\end{equation}

Finally, we define the slip indicator $\eta$ as whether either finger passes some max pre-defined threshold $T$. 
\[
\boldsymbol{\eta}=
\begin{cases}
1, & \text{if } \max_{g \in G} \bar{\sigma}_g > T,\\[1ex]
0, & \text{otherwise\,.}
\end{cases}
\]

We summarize the parameter values we use for \ourmethod{} in Tab.~\ref{tab:selected_parameters}. These parameters result in a slip detection frequency of 500~\si{\hertz} and a moving variance calculated with PSD features of the last 30~\si{\ms}. These parameters lead to FFT bins that span the frequency range $[7.5, 52.5]$~\si{\hertz}, approximating the range of mechanoreceptors in the human hand \cite{johansson2009codinghumanskinforobjectmanipulation}.

%% file: sections/characterization.tex
\begin{figure*}[t]
        \vspace{-15pt}
	\centering
	\includegraphics[width=\linewidth]{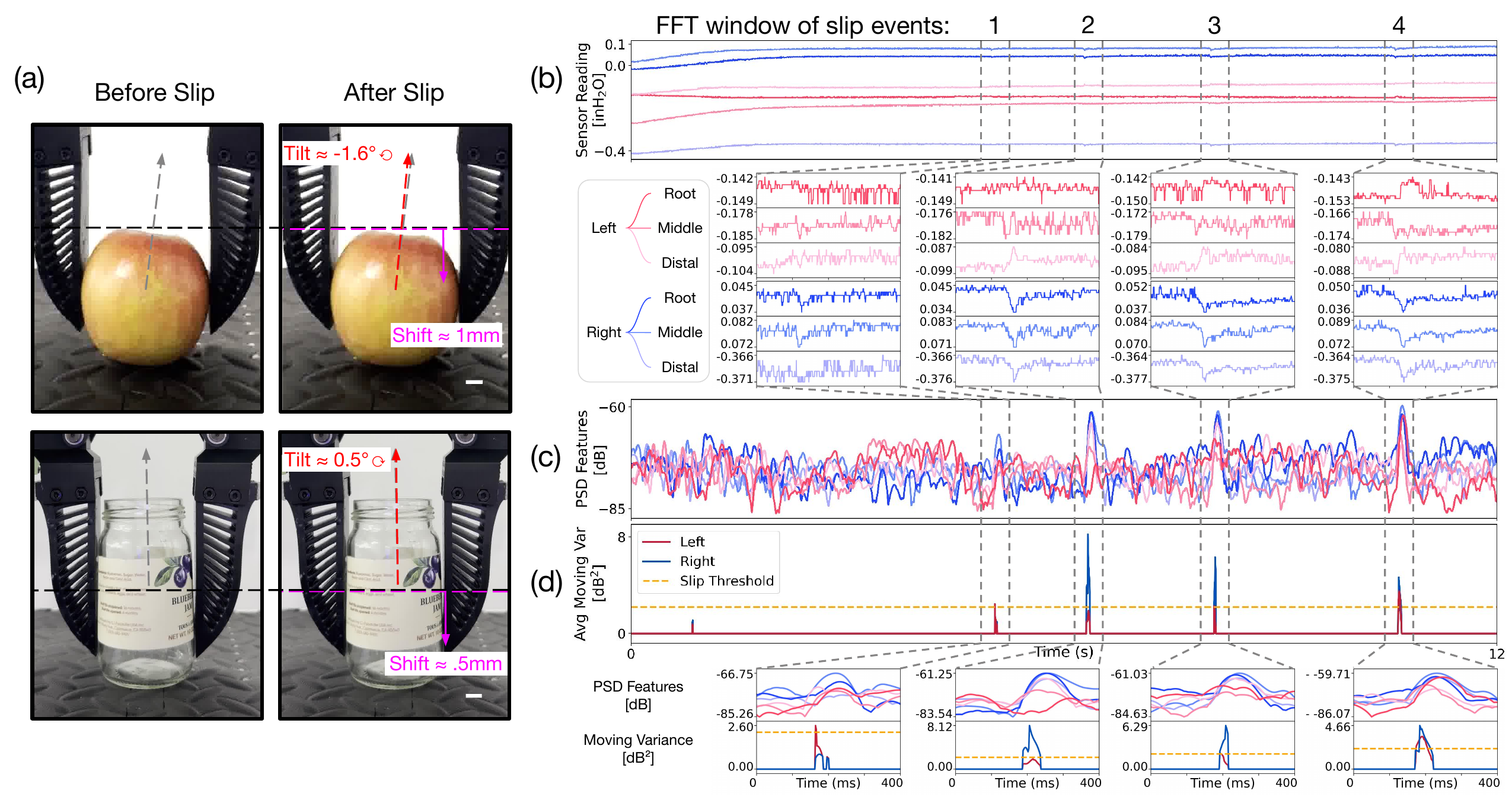}
        \vspace{-20pt}
	\caption
	{
        \textbf{Slip Detection Characterization} \textbf{(a)}  Snapshots of object displacements caused by stick-slip events. The time difference between the two snapshots of both objects is $66.7~\si{\milli\second}$. The snapshots of apple corresponds to Slip Event 1. Scale bars represent 1~\si{\cm}. \textbf{(b--d)} Filtered sensor readings, PSD features, and average moving variance of one representative trial of lifting an apple. There are four slip events during this trial. At the onset of each slip event, a sharp increase in the PSD feature is observed, resulting in a corresponding spike in the average moving variance, as further illustrated in the zoomed-in views.
        } 
        \vspace{-5pt}
    \label{fig:slip_characterization}
\end{figure*}

\section{Characterization}
\label{sec:characterization}

In this section, we characterize FORTE's ability to estimate grip force and detect slip during an attempted grasp. We attach the FORTE fin-ray fingers to a linkage-based gripper \cite{seo2024legato} and mount the entire  assembly to a Franka Emika Panda arm. We define $\theta_{\text{closed}} = 0\si{\degree}$ as the servo position when the gripper is fully closed. We define $\theta_{\text{zeroF}}$ as the servo position when the gripper's fingers \textit{just} touch the object, applying zero force. Each object has its own specific $\theta_{\text{zeroF}}$, depending on its size.

\subsection{Grip Force Estimation}
To characterize how well we measure grasp force across different shapes, we perform multiple grasps on custom load-cell testing indentors (Fig.~\ref{fig:force_characterization}). The load cell assembly is centered within the gripper's grasp, and a simple \textit{close gripper} operation is performed. For each of the 6 indentors, we collect data for 40 trials. In each trial, we close the gripper by a random amount, sampling uniformly from $\theta_{\text{closed}}$ to  $\theta_{\text{zeroF}}$. This creates a random grasping force against the load cell, which we compare against the FORTE readings. To process each trial, each channel's sensor readings are first normalized by subtracting the mean during initialization, then filtered using a median filter with a kernel size of 11. The resulted 3-hour force estimation dataset contains 108k data points.

Our filtered sensor signals correspond well to the grip force, as indicated by the 10-fold trial-wise cross-validation performance of our force estimation algorithm (Fig.~\ref{fig:force_characterization}). Our method achieved an average RMSE of 0.187 on a force range of 0 to 8~\si{\newton} across all indentor shapes. The estimation performance surpasses the error of 0.39~\si{\newton} over a 0.9–4.5~\si{\newton} range reported using a barometric strain gauge array~\cite{choi2023integratedbarosg}, and is comparable to the error of 0.17~\si{\newton} on a force range of 0.3 to 3.0~\si{\newton} using an external camera \cite{xu2021compliantexternal}. This represents a notable improvement over the baseline RMSE of 0.211 obtained without using the feature vector (Sec.~\ref{sec:f_e}), demonstrating the effectiveness of our approach in enhancing force estimation accuracy.

\subsection{Slip Detection}
\label{sec:slipCharacterization}
To characterize slip detection, we generate slip events by attempting to vertically lift an object from the tabletop with insufficient grasping force (Fig.~\ref{fig:slip_characterization}). 
%Our slip generation process created realistic slip events with delicate grip force from the tabletop for the parallel gripper. The only external forces are from gravity and contact with the table, thus requiring no additional constraints and human interference. 
We lift objects by controlling the Franka Emika Panda arm at a constant speed of 1mm/s at 20~\si{\hertz} using Cartesian-space velocity control with Deoxys \cite{zhu2022viola}. We selected an apple and a jam jar to evaluate slip detection. For each object, we define $\theta_{\text{grasp}}$, the servo position where the object is stably grasped. For the purposes of characterization, we determine $\theta_{\text{grasp}}$ by repeatedly lifting the gripper from a fixed pose while gradually decreasing the joint position from $\theta_{\text{\text{zeroF}}}$ until the object is securely lifted. For each trial, we sample a random gripper servo position $\hat{\theta} \in [\theta_{\text{zeroF}},\ \theta_{\text{grasp}})$. This will set the force applied by the gripper to be randomly less than the stable grasping force ($< \SI{0.5}{\newton}$), so we know the object will definitely slip. For the apple, $\hat{\theta} \in [\SI{0.552}{\radian}, \SI{0.527}{\radian}]$, while for the jam jar, $\hat{\theta} \in [\SI{0.476}{\radian}, \SI{0.460}{\radian}]$.
We then command the robot arm to lift the gripper at constant velocity. We record this motion and use pixel values from the video to estimate the object's rotational and translational displacements during slipping. We collect data for 5 trials for each object (Fig.~\ref{fig:slip_characterization}).

Upon analysis of the slip data, we first note that the two fingers report different slip results. This is expected as objects are not placed exactly at the midpoint of the gripper workspace, leading to uneven distribution of grasping forces. 

We name the finger with the larger contact force as the ``leading finger'' and the other as the ``trailing finger.'' Similarly, we name their corresponding contact force as ``leading force'' and ``trailing force.'' The information of the leading finger is revealed by the finger-wise moving variance feature as shown in Fig.~\ref{fig:slip_characterization}d. In the given example, the right finger is the leading finger in slip event 1, while the left finger becomes the leading finger in slip events 2 and 3. We note that in some cases, the magnitude of the trailing force is superseded by the table friction force. This is because the estimated grip force at $\theta_{\text{grasp}}$ (0.5~\si{\newton}) is significantly lower than the gravitational force of the apple (2.27~\si{\newton}) and the jam jar (1.42~\si{\newton}).
Therefore, we focus on detecting the slip events involving the leading finger, as these are the necessary conditions of gross slip (Section~\ref{sec:slipDefinition}).

We find that our slip detection algorithm can efficiently capture the slip events of transitions from incipient slip to gross slip with a fixed slip threshold $T = 2~\si{\decibel}^2$. From characterization, we identify two main types of slip events:
\begin{enumerate}
    \item \textbf{Leading-Finger Stick-Slip}: Stick-slip occurs at the leading finger while the trailing finger undergoes gross slip.
    \item \textbf{Simultaneous Stick-Slip}: Stick-slip occurs simultaneously at both fingers.
\end{enumerate}

Fig.~\ref{fig:slip_characterization} includes both types of slip events. Fig.~\ref{fig:slip_characterization}b shows the filtered sensor signals as input to our slip detection algorithm. The PSD features computed with Eq.~\eqref{equation:PSD} from the FFT windows of size 400 of the filtered sensor signals are shown in Fig~\ref{fig:slip_characterization}c. Fig.~\ref{fig:slip_characterization}d shows the finger-wise average moving variance as the result of Eq.~\eqref{eq:avgMV}. 
Comparing Fig.~\ref{fig:slip_characterization}c and Fig.~\ref{fig:slip_characterization}d illustrates the effectiveness of measuring the increase in power rather than the magnitude. To be more specific, the first slip event can barely be distinguished from the PSD feature values, while the moving variance cleanly captures the slip event as shown in Fig.~\ref{fig:slip_characterization}d. Zoomed-in views of the slip events indicate that the latency between the onset of sensor signal change and successful slip detection is consistently below 100~\si{\milli\second}. Overall, \textsc{FORTE} enables precise and real-time slip detection.

%% file: sections/results.tex
\section{Delicate Grasping Experiments}
\label{sec:exp_setup}
\begin{figure}[]
	\centering
	\includegraphics[width=\linewidth]{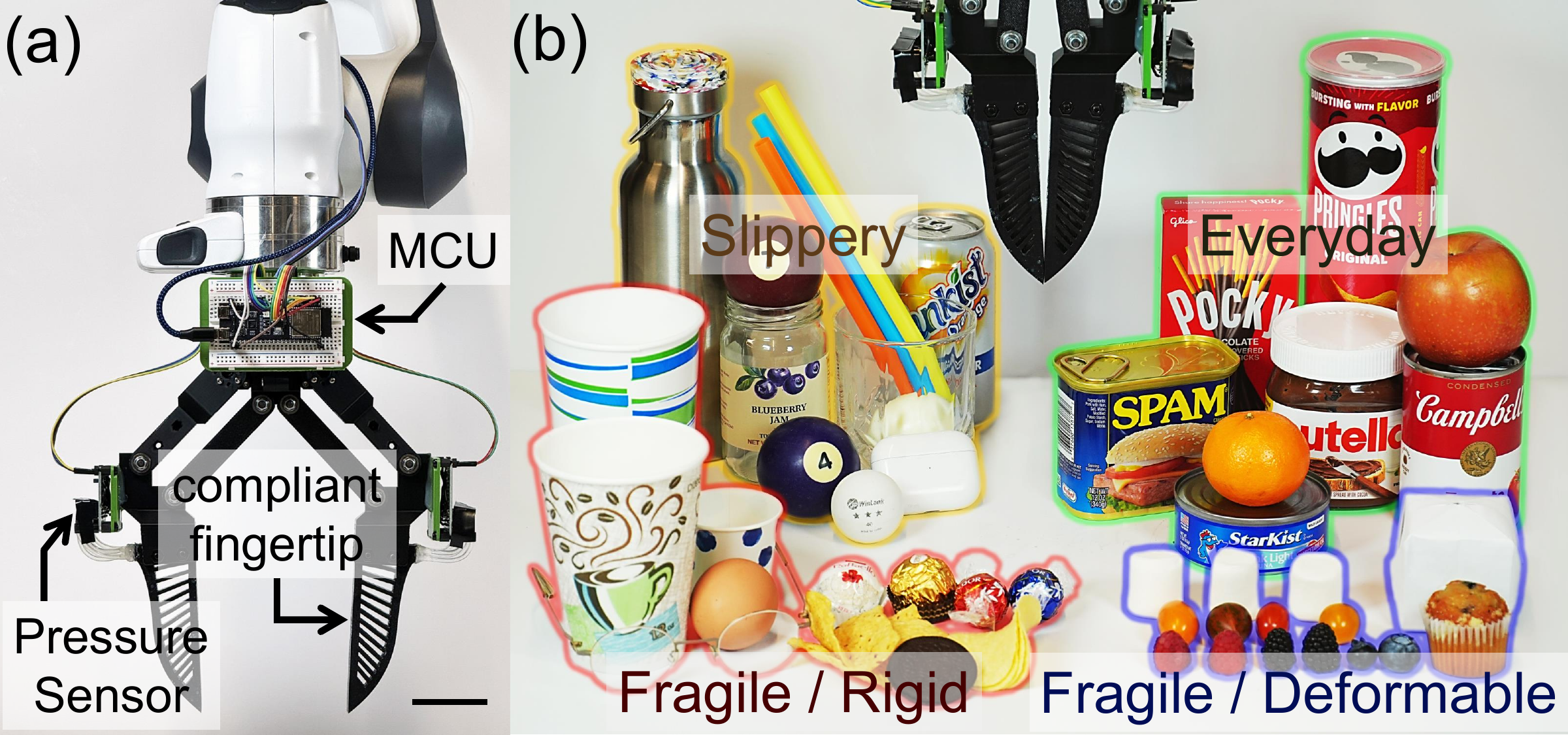}
        \vspace{-15pt}
	\caption
	{  \textbf{Experimental Setup (a)} \ourmethod{} is mounted on a Franka Emika Panda robot arm for table-top grasping tests. Scale bar represents 5~\si{\cm}.
        \textbf{(b)} Objects used for evaluation. Fragile objects are selected to evaluate force estimation accuracy and gripper compliance, while slippery items and eight everyday YCB objects are chosen to assess slip detection performance. 
        }
        \vspace{-12pt}
    \label{fig:objects}
\end{figure}

% Now that we have characterized FORTE's ability to estimate force and detect slip, we would like to know how well FORTE fares in real-world grasping scenarios. 
In this section, we evaluate \ourmethod{} against baselines in delicate object grasping. We created an evaluation set of 31 different object types (47 individual object instances), categorizing each object type into 4 categories: Fragile-rigid, Fragile-deformable, Slippery, and Everyday objects (Fig.~\ref{fig:objects}). The Everyday objects were selected from the YCB dataset \cite{calli2017ycb}. For each grasp strategy, we conducted 10 trials for each object type, reporting our results in Tab.~\ref{tab:result}. 

\input{tables/result_table}
\input{tables/conclusion_table}

\subsection{Experimental Setup}
% We evaluated FORTE and the baselines through a 31-item test set with 4 specific categories of interest: fragile and rigid objects, fragile and deformable objects, slippery objects, and everyday objects (Figure~\ref{fig:objects}). Items are drawn from real-world environments and the YCB food items \cite{calli2017ycb}. We conducted experiments of 10 independent trials for each of the objects on a set of 47 object instances across the 31 distinct object types (Table~\ref{tab:result}).

% in contact timing, gripping force, and lifting motion
For all grasping conditions, we used the same gripper setup from Sec.~\ref{sec:characterization} and adopted a standardized grasp procedure. We first initialize the gripper and manually place an object between its fingers. The gripper closes until contact is detected, defined as the maximum change in any tactile signal channel exceeding a threshold $\tau = 0.5\%$ of full-scale span within a sliding window of width $w = 200$ samples ($0.1~\si{\second}$). Upon contact, the gripper loads an initial force of $f_{\text{init}} = 0.25~\si{\newton}$ to ensure a consistent baseline. We then lift the object at $v_{\text{lift}} = \SI{5}{\milli\meter\per\second}$ using Cartesian velocity control at 20~\si{\hertz}. If slip is detected, the gripper command is incremented by $\Delta_{\text{cmd}} = \SI{0.88}{\degree}$ to increase the grip force. The lift continues until the object is either successfully lifted or dropped.

For \ourmethod{}, the force estimation module operates at 100~Hz with an update latency of less than 10~\si{\milli\second}, providing timely feedback for closed-loop control. The slip detection algorithm runs at 500~Hz as determined by the selected signal window size of 400 and an overlap factor of 0.99, and the typical detection latency is below 100~\si{\milli\second}. To evaluate the generalizability of our slip algorithms, we set the slip detection threshold $T$ by the values derived from the apple and jam jar used in Sec.~\ref{sec:slipCharacterization}. This means the majority of tested objects are unseen and out of distribution.

We compare \ourmethod{} against two baseline conditions:
\begin{itemize}
    \item \textbf{On-Off:} The gripper closes fully at a constant speed before lifting, without using any sensing. This baseline evaluates the performance of the compliant fingers alone.
    \item \textbf{W/o Slip:} Follows the standard procedure but disables slip detection. This baseline uses a fixed force without reactive adjustment to assess the role of slip sensing.
\end{itemize}

\subsection{Grasping Results}
Overall, \ourmethod{}'s force and slip sensing enabled a high success rate of 91.9\% among all the 310 trials, while On-Off and W/o Slip only achieved 60\% and 52.6\% success rates, respectively (Tab.~\ref{tab:result}). \ourmethod{}'s gentle initial gripping force of 0.25~\si{\newton} meant 13 out of 14 fragile objects were lifted with a 100 \% success rate. By contrast, the simple On-Off controller broke the surface of 8 out of the 14 selected objects and caused unrecoverable deformation of the blueberry, grape, tomato, and marshmallow. We do note that the On-Off controller had a 100\% success rate on the Raw Egg and the Chocolate Ball, highlighting that compliant fingers with a poor controller can still delicately manipulate some objects.

Meanwhile, while the W/o Slip Baseline performed on-par with \ourmethod{} in Fragile objects, it failed significantly in the Slippery and Everyday Objects category (31.1\% and 1.3\%). Indeed, 15 out of the 17 objects of the two categories were not able to be reliably lifted with the initial gripping force. The W/o Slip Baseline even performed worse than the On-Off baseline (88.8\% and 100\%). By contrast, \ourmethod{}'s use of adaptive grasping with slip detection resulted in a 100\% success rate for 10 out of the 15 objects and an overall category performance of 88.8\% and 85\%. We do note that the On-Off Baseline performed on par or better than \ourmethod{}, as it applies an excessively high force to prevent slippage. More fine tuning of the \ourmethod{}'s force controller may prevent that, at the expense of breaking more fragile objects.
% The sole failure that the W/o Slip controller had in lifting the Nutella jar is due to incorrect force prediction, caused by the object being placed too far off-center during initialization.

None of the three tested methods were able to lift hard-boiled eggs. FORTE and W/o Slip failed due to insufficient gripping force, while the On-Off baseline slowly squeezed the egg out of the gripper. Slippage was detected in all Pringles trials, but the object was dropped once when grasped sideways with the distal finger regions. The gripper failed to respond in time to the final slip, suggesting the need for improved control responsiveness for future applications.

In conclusion, FORTE demonstrates strong generalization across fragile, slippery, and everyday objects by combining a compliant finger with real-time force estimation and slip detection. These results underscore \ourmethod{}’s effectiveness in delicate manipulation, setting a solid foundation for more responsive robotic manipulation in real-world settings.

\subsection{Slip Detection Performance}
In addition to grasping performance, we also evaluated how accurate our slip detection algorithm is in these real-world scenarios. We recorded how many trials had slips during lifting and whether our algorithm made a true positive, true negative, or false positive prediction on slip. Summary statistics are reported in Tab.~\ref{tab:result}. When both ground truth slippage and predicted slippage are 0, the precision is defined as 1.0 to reflect the algorithm’s correct abstention from false positives.
 
The slip detection algorithm achieves an overall trial-wise accuracy of 0.93 and an F1 score of 0.91. The precision of our slip detection algorithm of 1 across all 310 trials demonstrates the trustworthiness of our slip detection algorithm. This high precision avoids unnecessary grasp adjustments and results in a high success rate of 98.6\% for grasping the fragile objects.

Since our slip detection algorithm targets the stick-slip events (Sec.~\ref{sec:slipCharacterization}), FORTE will primarily measure micro-vibrations at the contact surface caused by shifts from static to kinetic friction. This vibration measurement allows us to detect slip even in cases where there is not noticeable geometric changes, such as in long objects (ex. bottles, cans, jars). However, this measurement technique makes us less effective for objects with extremely smooth surfaces (ex. Pocky box, boiled egg). In these cases, the friction transition produces only minor changes in force, making it difficult for FORTE to detect.
    % typical failure cases arise when the applied gripping force is too low to generate detectable contact interactions between the finger surface and the object. 
    % The slip detector failed to detect the slippage of the hard-boiled egg and the Pocky box, since these two objects have a very low friction coefficient between the object surface and \ourmethod{}'s fingers. 
    Smooth and continuous gross slip initiates at the beginning of lifting, and the transition is not significant enough to generate a detectable power increase of PSD features. Similarly, among the 20 trials of grasping raw eggs and apples, our slip detection algorithm made 3 false negative predictions. In all three trials, the objects rolled out from the side of the gripper as rotational gross slip is initiated. This is primarily caused by the small contact area between the edge of the finger surface and the objects’ curved geometry, together with their off-centered center of mass.

In addition to the two types of targeted slip events defined in Sec.~\ref{sec:slipCharacterization}, we identified two other types of slip events that can be captured with our slip detection algorithm:
\begin{enumerate}
    \item \textbf{Rotational-to-Translational Slip Transition}: Slip begins as rotational gross slip and transitions into translational gross slip during lifting.
    \item \textbf{Texture-Affected Slip}: Local surface texture variations cause detectable fingertip deformation during gross slip.
\end{enumerate}

A typical case of rotational-to-translational slip transition includes side gripping at the off-centered position of objects of elongated geometries, such as the Pringles Bottle and the Metal Bottle. As one end of the bottle is lifted and the object tilts more gradually, translational gross slip will be initiated. A representative example of texture-affected slip is when slip is detected as the fingertip moves across the circumferential ridges along the side of a tomato soup can. Further work is needed to improve our algorithm to better capture these more differentiated types of slip events.

Overall, our slip detection algorithm demonstrates robust performance across unseen and diverse objects, with sufficient responsiveness to detect and react to slippage during lifting. Moreover, it captures multiple types of slip events through a unified analytical approach. This capability is essential for achieving delicate manipulation in real-world.

%% file: tables/result_table.tex
\begin{table*}[!t]
\centering
\renewcommand{\arraystretch}{0.95}
\caption{Category-level summary of grasp success rates and slip-detection performance across 31 object types.}
\vspace{-3pt}
\label{tab:result}
{
% \footnotesize
\begin{tabularx}{\linewidth}{lc*{3}{>{\centering}X}*{4}{c}}
\toprule
\multirow{2}{*}{\textbf{Category}} & \multirow{2}{*}{\textbf{\# Obj}} & 
\multicolumn{3}{c}{\textbf{Grasp Success Rate}} & 
\multicolumn{4}{c}{\textbf{Slip Detection Performance}} \\
\cmidrule(lr){3-5} \cmidrule(lr){6-9}
 & & On-Off & W/o Slip & FORTE (ours) & 
 Acc. & Prec. & Recall & F1 \\
\midrule
Fragile / Rigid      & 7  & 33.3\% (20/60)  & 91.4\% (64/70)  & \textbf{97.1\% (68/70)}  & 0.97 & 1.00 & 0.60 & 0.75 \\
Fragile / Deformable     & 7  & 0\% (0/70)      & \textbf{100\% (70/70)} & \textbf{100\% (70/70)} & 1.00 & 1.00 & 0.00 & 0.00 \\
Slippery     & 9  & \textbf{88.8\% (80/90)} & 31.1\% (28/90) & \textbf{88.8\% (80/90)} & 0.91 & 1.00 & 0.82 & 0.90 \\
Everyday     & 8  & \textbf{100\% (80/80)} & 1.3\% (1/80)  & 85.0\% (68/80)  & 0.88 & 1.00 & 0.86 & 0.93 \\
\midrule
\textbf{Overall} & 31 & 60.0\% (180/300) & 52.6\% (163/310) & \textbf{91.9\% (285/310)} & \textbf{0.93} & \textbf{1.00} & \textbf{0.84} & \textbf{0.91} \\
\bottomrule
\end{tabularx}
}
\vspace{-3pt}
\end{table*}

%% file: tables/conclusion_table.tex
\begin{table*}[!t]
\centering
\vspace{-3pt}
\caption{Comparison of Compliant Grippers with Integrated Force Estimation and Slip Detection}
\vspace{-3pt}
\label{tab:tactile_gripper_comparison}
% \scriptsize
\setlength{\tabcolsep}{5pt}
\renewcommand{\arraystretch}{0.97}

\begin{tabular}{@{}l l l c c c c@{}}
\hline
\textbf{Method} &
\mbox{\textbf{End-effector / Sensor Integration}} &
\textbf{Sensing Method} &
\textbf{Fs (Hz)} &
\mbox{\textbf{Force Est. Error}} &
\textbf{Lifespan (h)} &
\mbox{\textbf{Slip Lat. (ms)}} \\
\hline
FORTE (ours) &
Fin-ray + embedded air channels &
Barometric &
2k &
$\mathbf{0.19}$ N (0--8 N) &
$\mathbf{>200}$ &
$\mathbf{<100}$ \\

GelSight FinRay~\cite{liu2023gelsight} &
Fin-ray + elastomer membrane + camera &
Vision-based &
30--100 &
1.18 N (0--25 N) &
1--3~\cite{zhao2025polytouch} &
-- \\

Soft-Bubble~\cite{kuppuswamy2020softbubvispressure} &
Latex bubble surface + camera &
Vision-based &
30--60 &
1.24 N (0--15 N) &
-- &
-- \\

3D-ViTac~\cite{huang20243dCapFinray} &
Fin-ray + tactile array on surface &
Piezoresistive &
23 &
-- &
-- &
-- \\

PolyTouch~\cite{zhao2025polytouch} &
Elastomer membrane + camera + contact mic &
Vision + Acoustic &
30--48k &
-- &
35 &
-- \\

PneuFlex~\cite{wall2023pasacacousticsoft} &
Pneumatic actuator + embedded mic/speaker &
Acoustic &
1--24k &
98\% (3-bin cls.) &
-- &
-- \\
\hline
\end{tabular}
\vspace{-8pt}
\end{table*}

%% file: sections/conclusion.tex
\section{Discussion}

In this work, we introduced a system that integrates force and slip sensing into compliant robotic fingers with simple 3D-printing-based manufacturing. This combination enables a parallel gripper to perform delicate manipulation tasks, such as handling fragile objects like raspberries and potato chips. Leveraging accurate slip detection with a real-world accuracy of 0.91, our system achieves a single-trial grasping success rate of 91.9\% over 31 diverse objects. Over the course of development, the same pair of FORTE fingers has been used for more than 200 hours of active operation spanning six months, showing no noticeable degradation in structural integrity or sensing reliability. This stability indicates that the compliant geometry and sealed pneumatic channels can endure extensive contact cycles without fatigue or leakage. A quantitative comparison with compliant grippers with tactile sensing is summarized in Table~\ref{tab:tactile_gripper_comparison}. Notably, \ourmethod{} provides high-rate sensing at 2~kHz, enabling tactile feedback of both continuous deformation and transient slip events. This rich feedback supports accurate force estimation while allowing timely slip detection.

One notable limitation of \ourmethod{} is that the sensor signal is sensitive to temperature change on the finger surface. According to the ideal gas law \(PV = nRT\), a rise in temperature $T$ will only increase the internal pressure $P$ in the sealed air channels under the inner finger surface, and vice versa. This temperature-induced pressure variation introduces large-magnitude noise into the sensor readings and can cause the system to respond inconsistently to identical contact and slip events. This limitation can possibly be improved with a more advanced design of channel geometry and layout. Overall, \ourmethod{} presents a compelling solution for integrating force and slip sensing to compliant gripper fingers.

In the future, we aim to explore optimized material stiffness and geometry for specific gripper requirements. The stiffness of the elastomer can influence the sensor’s response characteristics as softer materials tend to provide higher sensitivity, whereas stiffer formulations enhance repeatability and durability. Moreover, increasing the friction coefficient by incorporating micro-texture on the finger surfaces could further improve slip detection performance, especially when handling objects with low-friction surfaces. Co-optimizing material properties with sensor layout will significantly enhance FORTE's performance in real-world applications.